\documentclass[12pt]{article}
\usepackage{marvosym}
\usepackage{amsmath} 
\usepackage{amsthm} 
\usepackage{amssymb}    
\usepackage{graphicx} 
\usepackage[dvips,letterpaper,margin=1in,bottom=0.7in]{geometry}
\usepackage{tensor}
\usepackage{calc}
\usepackage{tikz}

\usepackage{amsmath}
\makeatletter 
\renewcommand{\maketitle} 
{ \begingroup \vskip 10pt \begin{center} \huge {\bf \@title}
    \vskip 10pt \large \@author \hskip 20pt \@date \end{center}
  \vskip 10pt \endgroup \setcounter{footnote}{0} }
\makeatother 
\usepackage{enumerate}
 
\let\baraccent=\= 
\renewcommand{\=}[1]{\stackrel{#1}{=}} 

\providecommand{\bd}[1]{\textbf{#1}}
\providecommand{\sct}[1]{\section{#1}}
\providecommand{\sbs}[1]{\subsection{#1}}

\providecommand{\be}[1]{\begin{enumerate}[#1]}

\providecommand{\ee}{\end{enumerate}}
\providecommand{\bq}{\begin{quote}}
\providecommand{\eq}{\end{quote}}

\usepackage{cancel}

\theoremstyle{definition}

\newbox\gnBoxA
\newdimen\gnCornerHgt
\setbox\gnBoxA=\hbox{$\ulcorner$}
\global\gnCornerHgt=\ht\gnBoxA
\newdimen\gnArgHgt
\def\gd #1{%
\setbox\gnBoxA=\hbox{$#1$}%
\gnArgHgt=\ht\gnBoxA%
\ifnum     \gnArgHgt<\gnCornerHgt \gnArgHgt=0pt%
\else \advance \gnArgHgt by -\gnCornerHgt%
\fi \raise\gnArgHgt\hbox{$\ulcorner$} \box\gnBoxA %
\raise\gnArgHgt\hbox{$\urcorner$}}

\providecommand{\beg}{\begin{enumerate}[I]}
\providecommand{\bee}{\begin{enumerate}[i]}

\providecommand{\ee}{\end{enumerate}}

\usepackage{forest}

\title{Cognitive bias in large language models: Cautious optimism meets anti-Panglossian meliorism}
\author{David Thorstad $|$ Vanderbilt University}
\date{}

\usepackage{txfonts}
\usepackage{pxfonts}
\usepackage{url}
\usepackage{booktabs}
\usepackage{wrapfig}

\newcounter{numbcounter}
\makeatletter
\renewcommand\@biblabel[1]{}

\makeatother

\usepackage{setspace}

\usepackage{natbib}
\setcitestyle{aysep={}} 

\usepackage{float}
\usetikzlibrary{trees}

\usepackage{setspace}

\begin{document}

\maketitle

\begin{abstract} \noindent Traditional discussions of bias in large language models focus on a conception of bias closely tied to unfairness, especially as affecting marginalized groups. Recent work raises the novel possibility of assessing the outputs of large language models for a range of cognitive biases familiar from research in judgment and decisionmaking. My aim in this paper is to draw two lessons from recent discussions of cognitive bias in large language models: cautious optimism about the prevalence of bias in current models coupled with an anti-Panglossian willingness to concede the existence of some genuine biases and work to reduce them. I draw out philosophical implications of this discussion for the rationality of human cognitive biases as well as the role of unrepresentative data in driving model biases. \end{abstract}

\doublespacing

\sct{Introduction}

The recent success of large language models gives new urgency to the question of how model performance should be evaluated. In many tasks, models can be evaluated for the accuracy of their outputs. However, models can also be evaluated along other important dimensions. For example, we can assess models for the transparency or interpretability of their judgments \citep{Creel2020,Vredenburgh2022}. We can also assess models for the presence of problematic biases \citep{Kelly2023,Johnson2020}.

Most work on biases in large language models focuses on a conception of bias closely tied to unfairness, especially as affecting marginalized social groups. However, recent work has alleged that large language models also show a number of classic cognitive biases familiar from work in the psychology of reasoning, behavioral economics, and judgment and decisionmaking \citep{Dasgupta2022,Lin-Ng-2023,Jones-Steinhardt-2022}. 

This development is exciting because it raises the possibility of using cognitive bias as a novel metric by which to evaluate the performance of large language models. A natural question to ask is how well existing systems perform along the metric of cognitive bias. By contrast to recent work on algorithmic bias, my aim in this paper is to offer a qualified piece of good news: existing evidence does not support the attribution of widespread and problematic cognitive biases to large language models. 

In more detail, my aim in this paper is to draw two lessons from recent discussions of cognitive bias in large language models. The first lesson is cautious optimism about model performance. In particular, many studies find biases which have standard rationalizing explanations when produced by humans. I argue that these explanations often generalize to show that the claimed biases are desirable features of reasoning by large language models (Section \ref{KnowledgeEffects}), in the process reinforcing the robustness of standard rationalizing explanations in the human case by showing how similar cognitive phenomena arise in agents with highly distinct cognitive architectures \citep{Dasgupta2022}. Furthermore, some studies find especially benign forms of classic biases (Sections \ref{Availability}-\ref{Anchoring}), whose desirability is particularly difficult to contest.

The second lesson is an anti-Panglossian willingness to accept the existence of some genuine and undesirable cognitive biases in reasoning by existing large language models. In particular, I argue that many models show framing effects (Section \ref{Framing}) and that these effects are not always desirable. When faced with undesirable biases, I argue that the proper reaction is to work to mitigate the bias, but not to exaggerate the prevalence or undesirability of biases in assessing overall model performance.

Here is the plan. Section \ref{Preliminaries} begins with two preliminary remarks. Sections \ref{KnowledgeEffects}-\ref{Anchoring} then make the case for cautious optimism through case studies of knowledge effects (Section \ref{KnowledgeEffects}), availability bias (Section \ref{Availability}) and anchoring bias (Section \ref{Anchoring}). Section \ref{Framing} makes the case for an anti-Panglossian willingness to accept at least one problematic bias: framing effects. Section \ref{TwoLessons} uses these discussions to elaborate and justify the reactions of cautious optimism and anti-Panglossian meliorism. Section \ref{Conclusion} concludes by drawing philosophical implications concerning the role of unrepresentative data in producing model biases (Section \ref{UnrepresentativeData}) and the rationality of biases in human cognition (Section \ref{VindicatoryEpistemology}).

\sct{Preliminaries} \label{Preliminaries}

Before beginning, two remarks are in order. First, as Richard Shiffrin and Melanie Mitchell \citeyearpar{Shiffrin2023} remind us, it is important to avoid inappropriate anthropomorphism in describing the performance of large language models. Some theorists may be comfortable using anthropomorphic vocabulary in which models are described as reasoning to judgments, which can be rational or irrational. Others will prefer a more neutral paraphrase, in which models are described as returning outputs in response to prompts, where the outputs may be desirable or undesirable given users' goals. I will sometimes use cognitive vocabulary, such as reasoning and judging, to describe model performance, although readers are welcome to substitute their preferred de-anthropomorphized paraphrase. On the other hand, I will not describe model outputs as rational or irrational, but only as desirable or undesirable. This reflects a lack of commitment to the judgments made by large language models having normative status in their own right. This contrasts with the case of human judgment, where it makes sense not only to describe biases as rational or irrational, but also to ask (Section \ref{VindicatoryEpistemology}) how the study of biases in language models bears on the rationality of biases in human cognition.

Second, recent findings suggest that patterns of bias in large language models may be highly model-sensitive. For example, Thilo Hagendorff and colleagues \citeyearpar{Hagendorff2023} find atypical performance by GPT-1 and GPT-2 in reasoning tasks, human-like performance by GPT-3, and hyperrational performance by GPT-4. Likewise, John J. Horton \citeyearpar{Horton2023} finds atypical behavior by models prior to GPT-3, but humanlike behavior in GPT-3. Given these findings, it is very important to specify the model used in each finding, which I will do in all cases where a finding is extensively discussed. 

There does remain some danger that the discussion in this paper will be superseded or rendered moot by further technological changes, leading to changes in patterns of model reasoning. This is a risk faced by a great deal of research in the philosophy of artificial intelligence, and it is a risk that must be openly admitted without dissembling. 

With these remarks in order, the next order of business is to look at four types of biases that have been alleged in large language models: knowledge effects (Section \ref{KnowledgeEffects}), availability bias (Section \ref{Availability}), anchoring bias (Section \ref{Anchoring}) and framing effects (Section \ref{Framing}). I will suggest that the first three findings may not be undesirable, but that some framing effects are probably undesirable and should be mitigated.

\sct{Knowledge effects} \label{KnowledgeEffects}
 
\sbs{Background} \label{KnowledgeBackground}

For much of the twentieth century, human reasoning was understood using a logical paradigm \citep{Wason1968,Rips1994}. Agents asked to assess the quality of inferences were assumed to test them for logical validity. Conditional claims were modeled using the material conditional, and conditional rules were to be tested by trying to falsify the embodied material conditional.

A probabilistic turn throughout the academy \citep{Erk2022,Ghahramani2015} has come to psychology \citep{Chater2006}, and in particular to the psychology of reasoning. There, `new paradigm' Bayesian approaches suggest that humans often do and should interpret reasoning tasks probabilistically, rather than logically \citep{Elqayam2013,Oaksford2007}. On Bayesian approaches, conditional assertions are licensed if the consequent has high probability conditional on the antecedent \citep{Oaksford2007}; conditional rules are tested by reducing uncertainty about the probabilistic dependency between consequent and antecedent \citep{Oaksford1994}; and inferences are tested for probabilistic forms of validity \citep{Adams1975}.

Logical and probabilistic paradigms come apart in their treatment of \textit{knowledge effects}: the influence of prior knowledge on reasoning in ways not licensed by classical logic. For example, agents are more likely to endorse an inference if they are more confident in its conclusion. On a logical paradigm, this finding was taken to reflect a problematic \textit{belief bias} to judge arguments with believed conclusions to be logically valid \citep{Evans1983}. But on a probabilistic paradigm, this finding is to be expected: good inferences should secure high-probability conclusions, and the prior probability of a conclusion has an important effect on its probability at the end of an inference \citep{Adams1975,Oaksford2007}.

Many large language models show human-like knowledge effects in a variety of tasks, including the Wason selection task \citep{Binz2023} as well as syllogistic and natural-language reasoning problems \citep{Dasgupta2022}. In this section, I introduce one salient knowledge effect (Section \ref{WasonSelection}) then argue that the effect should be viewed at least as favorably in large language models as it is viewed in humans (Section \ref{FeatureBug}).

\sbs{Wason selection} \label{WasonSelection}

Suppose you are shown four two-sided cards. Their visible sides contain an ace, king, two and seven, respectively (Figure \ref{WasonFig}). You are asked to test the rule that `If a card has an ace on one side, then it has a two on the other'. Which cards should you turn over to test the rule?

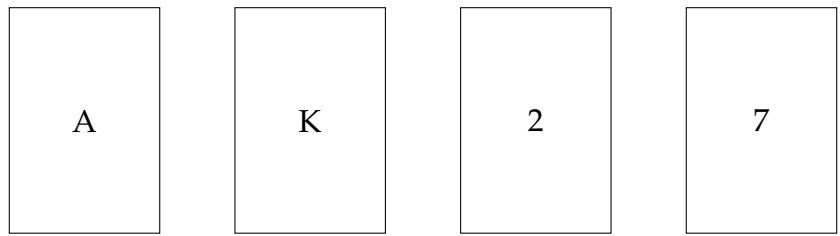
\begin{figure}[!h]
\begin{center}
\begin{tikzpicture}
  \draw (0, 0) rectangle (2, 3);
  \node at (1, 1.5) {A};
  
  \draw (3, 0) rectangle (5, 3);
  \node at (4, 1.5) {K};
  
  \draw (6, 0) rectangle (8, 3);
  \node at (7, 1.5) {2};
  
  \draw (9, 0) rectangle (11, 3);
  \node at (10, 1.5) {7};
\end{tikzpicture}
\end{center}
\caption{The Wason selection task}
\label{WasonFig}
\end{figure}

Let us label the cards as $p$ (A), $\lnot p$ (K), $q$ ($2$) and $\lnot q$ ($7$). In this notation, the rule is `If $p$, then $q$'. On a logical interpretation, the rule expresses the material conditional $p \supset q$, which is tested by searching for falsifying instances $p \land \lnot q$. This means that agents should turn the $p$ and $\lnot q$ cards, that is the ace and the seven. Wason's original finding, replicated across countless subsequent experiments, is that far fewer than ten percent of agents make the logically correct choice \citep{Wason1968}. 

This behavior is poor enough for such a simple task that we are well within our rights to ask whether agents might have interpreted the task probabilistically rather than logically. The classic Bayesian approach to the Wason selection task is due to Mike Oaksford and Nick Chater \citeyearpar{Oaksford1994}.

On this approach, agents turn cards in order to reduce uncertainty about the probabilistic relationship between the propositions $p$ and $q$ expressed in the conditional rule. On the simplest model, they want to discriminate between two hypotheses: the \textit{dependence hypothesis} $P(q|p) = 1$ that $p$ and $q$ are probabilistically dependent, and the \textit{independence hypothesis} $P(q|p) = P(q)$ that $p$ and $q$ are probabilistically independent. 

Oaksford and Chater make two additional assumptions. First, they assume that the uncertainty which agents aim to reduce is measured by Shannon entropy \citep{Shannon1948}.\footnote{The Shannon entropy of credence function $P$ is $\Sigma_{X = \{M_I, M_D\}} P(X)log_2(P(X))$. This definition enforces Oaksford and Chater's assumption that the agent has beliefs about the independence hypothesis $M_I$ and dependence hypothesis $M_D$ and aims to reduce her uncertainty about these hypotheses.} This is a common assumption drawn from research in information theory. Second, Oaksford and Chater assume that agents treat $p$ and $q$ as somewhat antecedently implausible. This is justified by research suggesting that agents do and should treat most propositions as improbable in causal reasoning, due to factors such as the large number of possible alternatives \citep{Anderson1990}. That assumption places us within the realm of knowledge effects: manipulations to increase the prior probability of $p$ and $q$ change Wason selection behavior \citep{Oaksford1994}. 

Under these assumptions, we can show that uncertainty reduction is maximized by turning the $p$ and $q$ cards, that is the ace and the two. And that is just what agents tend to do \citep{Oaksford1994}. In this way, the Oaksford and Chater model provides a probabilistic explanation for why agents do, and perhaps should, turn the cards that they choose to turn. 

Ishita Dasgupta and colleagues \citeyearpar{Dasgupta2022} test the Chinchilla model \citep{Hoffman2022} on several versions of the Wason selection task. They find across task versions that the model is no more than about 50\% likely to take the logically correct action of turning the $p$ and $\lnot q$ cards, and in many conditions the model is at most 20\% likely to do so (Figure \ref{WasonFig}). In particular, Dasgupta and colleagues find a significant tendency to turn the $q$ card. As Dasgupta and colleagues note, these patterns of behavior conform in coarse outline to the predictions of Oaksford and Chater's probabilistic model but conform less well to the logical model. 

\begin{figure}
\begin{center}
\includegraphics[scale=0.5]{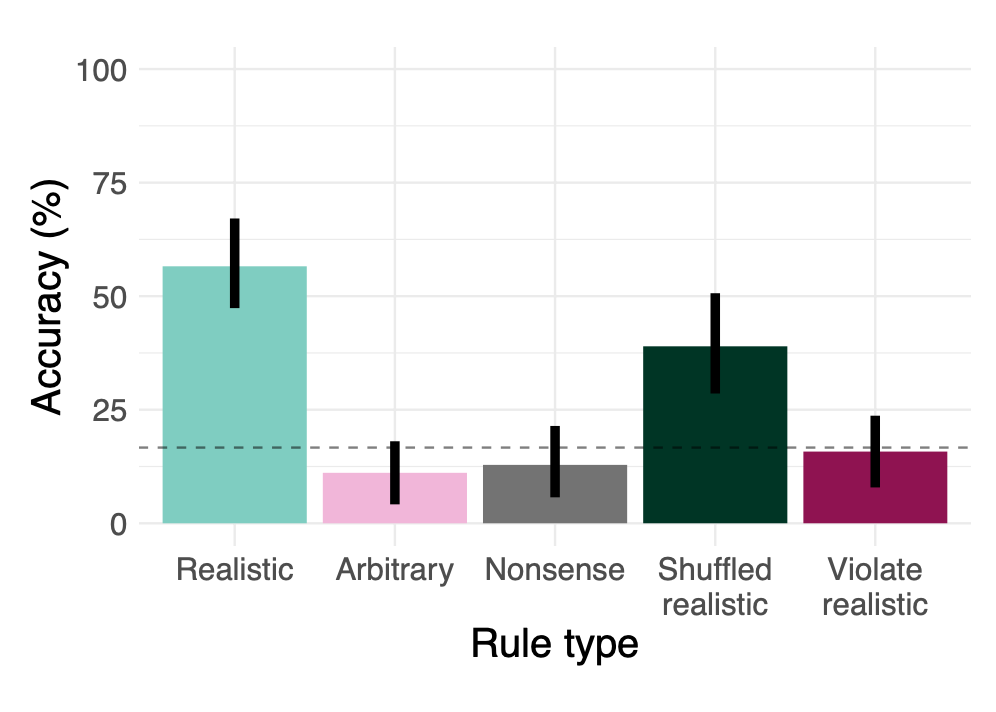}
\caption{Wason selection task performance (logical criterion) by Chinchilla across rule types, from \citet{Dasgupta2022}.}
\label{WasonFig}
\end{center}
\end{figure}

\sbs{A feature or a bug?} \label{FeatureBug}

Should knowledge effects be treated as a desirable feature of large language models, or an undesirable bug to be driven out of them? To a large extent, I think that we should answer this question in the same way as we answer it for humans. Those sympathetic to Bayesian approaches stress that while logic is well-suited for reasoning under certainty, probabilistic approaches are well-suited for reasoning in an increasingly uncertain and data-driven world. Probabilistic approaches view knowledge effects as desirable uses of prior knowledge to improve reasoning. Those sympathetic to logical approaches will no doubt disagree, but this is not the place to re-litigate ongoing normative disputes between the logical and probabilistic paradigms.

However, there may be two reasons to look more favorably on knowledge effects in large language models than in humans. The first is that previous logical paradigms in artificial intelligence have been challenged by increasingly successful probabilistic approaches \citep{Ghahramani2015}. It is now thought that probabilistic systems often outperform logic-based agents in the data-laden, uncertainty-rich contexts which large language models confront: exactly the conditions under which probabilists suggested they should. If this is right, then even if we think that humans often do better to reason logically, we needn't enforce the same constraint on deep learning agents, who are increasingly successful in combining probabilistic tools with data to make sense of the world.

Second, there is good evidence that many large language models can learn the logical interpretations of reasoning tasks when they are asked to. For example, Dasgupta and colleagues also find that the Chinchilla model learns after just five training instances to nearly eliminate belief bias in natural language inference, and shifts substantially towards logical performance in the Wason selection task \citep{Dasgupta2022}.\footnote{Interpreting Wason selection task data is difficult because Dasgupta and colleagues find less movement towards the logical interpretation with non-realistic prompts. It is well known that humans also react quite differently to realistic versions of the Wason selection task than to non-realistic versions. What to make of this finding in human reasoning is an active area of descriptive and normative dispute \citep{Cheng1985,Cosmides1989,Oaksford1994}, and the same disputes may transfer to the machine case as well.} This suggests that if probabilistic construals of reasoning tasks are a feature of many large language models, they are not a deep feature ingrained by limits in cognitive abilities, as some authors have suggested that they are in the human case \citep{Evans2003}. Instead, large language models often retain the ability to reason either logically or probabilistically, and inducing logical reasoning may be as simple as telling the models that we would like them to reason logically.

\sct{Availability} \label{Availability}

If we are going to find uncontroversially problematic cognitive biases in large language models, we will need to look beyond knowledge effects. A natural place to start is by replicating classic biases from the heuristics and biases paradigm. In this section and the next, I explore attempts to find two of the three original biases proposed within this paradigm: availability bias and anchoring bias. I suggest that both attempts encounter significant obstacles, revealing important descriptive and normative lessons for future study.

\sbs{Current research on availability} \label{CurrentResearchAvailability}

In the early 1970s, Daniel Kahneman and Amos Tversky proposed that humans often make inferences using the \textit{availability heuristic} of ``estimat[ing] frequency or probability by the ease with which instances or associations could be brought to mind'' \citep[p. 208]{Tversky1973}. For example, participants presented with a list of 19 famous female actors and 20 less-famous male actors subsequently recalled the list as containing more female than male actors \citep{Tversky1973}. A natural explanation for this finding invokes availability: because participants were more readily able to bring female actors to mind during subsequent recall, they judged that the list contained more female than male actors.

It is now almost universally acknowledged that early discussions of the availability heuristic passed too freely between two senses of availability \citep{Schwartz2002}. \textit{Subjective availability} involves reliance on features of the subjective experience of the recall process, such as the felt ease or fluency with which information comes to mind. In this sense, agents may judge male actors to be rare if they strain and feel disfluency in trying to recall male actors. By contrast, \textit{objective availability} involves reliance on the content of information retrieved, or on non-experiential features of the retrieval process such as the time needed to retrieve information. In this sense, agents may judge male actors to be rare if they cannot recall many male actors, or if it takes a long time to recall male actors.

Few theorists hold that reliance on objective availability of information is always irrational or undesirable. If we can quickly bring many examples of a category to mind, then that provides some evidence that the category is common in our experience, and hence in the world. This much is conceded by Tversky and Kahneman themselves.\footnote{``Availability is an ecologically valid clue for the judgment of frequency because, in general, frequent events are easier to recall or imagine than infrequent ones'' \citep[p. 209]{Tversky1973}} Of course, to say that reliance on objective availability is sometimes desirable is not to say that uncritical deference to objective availability is desirable. Objective availability may be skewed by task-irrelevant factors such as the fame of actors, and agents must take appropriate steps to correct for these biasing factors. But no theory of human rationality or desirable model performance should fix a target of complete unreliance on objective availability.  

Matters are more complicated with regard to subjective availability. For present purposes, it is enough to say that subjective availability is not at issue in assessing current large language models, since it has not been alleged that large language models rely on, or even have such a thing as a subjective experience of memory retrieval. The irrationality or undesirability of subjective availability has been challenged in recent areas such as metacognition, where detailed and nuanced patterns of reliance on subjective feelings of fluency are thought to explain much of the success of human metacognition \citep{Alter2009,Proust2013}. However, for present purposes we may restrict attention to objective availability. 

\sbs{Availability in relation extraction}

Relation extraction tasks involve identifying relationships between objects from textual discussions of those objects. A paradigmatic relationship extraction task is the task of identifying drug-drug interactions \citep{Zhang2020}. Given a textual description of the interaction between two drugs, the algorithm must classify the type of interaction between them.

The Drug-Drug Interaction (DDI) dataset is an annotated corpus of 1,017 texts describing 5,021 interactions between various drugs \citep{Segura-Bedmar-2013}. Each discussion is annotated with one of five interaction types: \textit{mechanism} for a description of the interaction mechanism; \textit{effect} for a description of the effect itself; \textit{advice} for recommendations about how to respond to drug-drug interactions; \textit{int} for nonspecific descriptions of interactions, and \textit{negative} for non-interactions. The vast majority (85.2\%) of interactions in the DDI dataset are negative, and models trained on the DDI dataset understandably learn to reflect this fact.

Ruixi Lin and Hwee Tou Ng \citeyearpar{Lin-Ng-2023} train GPT-3 on the DDI dataset. Lin and Ng then test the model on `content-free' descriptions generated from the DDI dataset by replacing all medical terms with the dummy descriptor `N/A'. Lin and Ng propose that because the model has no direct knowledge of the dummy class, the model should classify dummy sentences according to a uniform probability description. That is, it should be 20\% likely to assign dummy descriptions to each interaction type: mechanism, effect, advice, int, and negative. 

Lin and Ng propose that any deviation from the uniform classification of dummy sentences should be treated as a form of availability bias, in which model judgments are skewed by the availability of interaction types in the training data. For each interaction type, Lin and Ng define the \textit{availability bias score} of that interaction type to be the absolute difference between the percentage of test items classified under this type and the 20\% classification rate expected under a uniform model. Under this definition, Lin and Ng find a strong availability bias, increasing in the number of descriptions used to train the model (Table \ref{DDI-Table}).

\begin{table}
\begin{center}

\begin{tabular}{@{}llllll@{}}
\toprule
\textbf{Training Examples}                                                                          & 10   & 100  & 1,000 & 10,000 & 25,296 \\ \midrule
\textbf{\begin{tabular}[c]{@{}l@{}}Availability Bias Towards\\ Negative Category (\%)\end{tabular}} & 26.3 & 77.7 & 39.7  & 47.0   & 52.0  \\ \bottomrule
\end{tabular}

\end{center}
\caption{Availability bias in drug-drug interaction by size of training set, Lin and Ng \citeyearpar{Lin-Ng-2023}.}
\label{DDI-Table}
\end{table}

Section \ref{CurrentResearchAvailability} distinguished between two forms of availability: objective and subjective. Lin and Ng's experiment studies a form of objective availability: the content of information stored in training data. This is an especially benign form of objective availability, because we are concerned with the availability of \textit{information} rather than with experiential properties of the information retrieval process, and we are concerned with the \textit{total} information stored in memory rather than a potentially unrepresentative sample retrieved during decisionmaking. Section \ref{CurrentResearchAvailability} suggested that many instances of objective availability should be regarded as unproblematic, and that seems a natural approach to the results presented by Lin and Ng.

Lin and Ng hold that because the model has no specific information about the dummy descriptor `N/A', ``the best that an unbiased model can do is to make a uniform random guess'' \citep{Lin-Ng-2023}. Traditional results in Bayesian epistemology suggest otherwise. Training on the DDI dataset provides the model with valuable information about the distribution of drug-drug interaction types across drugs. Rational Bayesian inference involves combining this prior information with novel information provided by descriptions to determine the probability that each given interaction is at play. Since the model has been exposed to primarily negative interactions during training, the model correctly learns that negative interactions are more common than positive interactions and learns to project this relationship onto novel drugs. When the model is exposed to larger samples of training data, it becomes more confident that negative interactions are prevalent. In the absence of competing information to move the model away from the prior, priors dominate and the model shows a strong tendency to predict novel drug-drug interactions to be negative, increasing in the quantity of training data. From an orthodox Bayesian standpoint, this is desirable behavior that should not be driven out of classification models. If anything, Lin and Ng's data show under-reliance, rather than over-reliance, on prior knowledge of interaction types.

Lin and Ng do suggest one more plausible lesson from this discussion: labels matter. While many machine learning scientists expect label information to become unimportant after training, testing models on content-free sentences reminds us of the importance of labels, since these sentences will be more likely to be classified using labels that are more frequent in the training data.\footnote{Tony Zhao and colleagues \citeyearpar{Zhao2021} call this majority-label bias.} However, it is not clear that forcing a uniform distribution of classification on content-free sentences is the right way to reduce the influence of arbitrary labels. After all, there is considerable arbitrariness in the number of labels used in the training data: for example, we could easily imagine the positive interactions being collapsed under a single label instead of four. Under a uniform distribution, this would increase the probability of negative predictions from 20\% to 50\%, a type of label-sensitivity that more traditional Bayesian methods avoid. 

One further lesson from this discussion is the importance of ecologically valid training data \citep{Todd2012b}. Models need to be exposed to data that is representative of the phenomena they will encounter during test, so that they will know how to predict the target phenomenon and not be distracted by distortions in the training data. This much is familiar from recent discussions of algorithmic fairness \citep{Hedden2021,Johnson2023}. Perhaps Lin and Ng's suggestion is that the DDI dataset is unrepresentative in its high proportion of negative drug-drug interactions, and if that is the case they will certainly have a point. However, if that is true, this failure should not be blamed on classifier algorithms. It should instead be blamed on those who collect and generate ecologically invalid datasets, or who use those datasets to train models to perform tasks for which the training data will no longer be representative. 

\sct{Heuristics and biases: Anchoring} \label{Anchoring}

\sbs{Current research on anchoring} \label{ResearchAnchoring}

The second of Tversky and Kahneman's original three heuristics is \textit{anchoring and adjustment} \citep{Tversky1974}. Suppose I ask you to estimate the year in which George Washington was first elected president. You might answer by \textit{anchoring} on an initial quantity, the year (1776) in which the Revolutionary War began, then \textit{adjusting} upwards and downwards to incorporate relevant knowledge, such as the length of the Revolutionary War and the drafting of the Constitution. If you are like most people, you might settle on an estimate around $1786.5$ \citep{Lieder2018}, which is quite good: Washington was elected in $1789$.

As this example illustrates, anchoring and adjustment produces a characteristic \textit{anchoring effect} in which judgments are skewed toward the initial anchor. $1786.5$ is quite close to the correct answer, but biased downwards towards the low anchor of $1776$. Anchoring effects are traditionally explained as the result of insufficient adjustments away from the initial anchor.

Tversky and Kahneman \citeyearpar{Tversky1974} initially proposed that a great number of anchoring effects should be explained as the result of mental processes of anchoring and adjustment. For example, Tversky and Kahneman instructed participants to spin a wheel, then judge whether the number displayed on the wheel was higher or lower than the number of African countries in the United Nations, and finally to estimate the number of African countries in the United Nations. Tversky and Kahneman found that judgments tended to be biased toward the value displayed on the wheel. Tversky and Kahneman explained this finding by assuming that agents anchored on an initial belief that the number of African countries in the United Nations is equal to the value on the wheel, then iteratively adjusted away from the anchor through a process of anchoring and adjustment.

That is a surprisingly irrational cognitive process, and subsequent authors rightly asked for evidence that a process of iterative anchoring and adjustment had in fact been employed. For two decades, all available process-tracing studies showed no evidence of a cognitive process of anchoring and adjustment in this and other early experiments \citep{Johnson1989,Lopes1982}. More recently, evidence has emerged that a genuine process of anchoring and adjustment may be employed in a small number of examples, such as our initial example of estimating the year in which George Washington was first elected president \citep{Epley2006,Lieder2018}. However, it is widely agreed that genuine anchoring and adjustment is extremely rare; that anchoring and adjustment is not typically triggered by external manipulations such as spinning wheels; that anchors tend to be relevant and informative, and incorporated in a rational way; that the results of anchoring and adjustment are often highly reliable; and that few if any anchoring effects in the early literature are produced by genuine processes of anchoring and adjustment \citep{Lieder2018}.

As evidence for processes of anchoring and adjustment failed to materialize in the motivating examples, researchers broadened the concept of anchoring effects so that they were no longer conceptually tied to a process of anchoring and adjustment. This broadening led to some confusion over the definition of anchoring effects, as Kahneman himself remarks:

\begin{quote}
The terms \textit{anchor} and \textit{anchoring effect} have been used in the psychological literature to cover a bewildering array of diverse experimental manipulations and results $\dots$ The proliferation of meanings is a serious hindrance to theoretical progress. \citep[p. 1161]{Jacowitz1995}.
\end{quote}

\noindent Many theorists outside the heuristics and biases camp have taken the definitional vagueness of biases such as anchoring as a mark against attempts to posit them \citep{Gigerenzer1996b}. For my part, I have some sympathy for this line, but I am willing to ask what anchoring effects might mean.

Here is a sampling of recent definitions of anchoring effects:

\begin{quote}
An anchor is an arbitrary value that the subject is caused to consider before making a numerical estimate. An anchoring effect is demonstrated by showing that the estimates of groups shown different anchors tend to remain close to those anchors. \citep[p. 1161]{Jacowitz1995}.
\end{quote}

\begin{quote}
The anchoring effect is the disproportionate influence on decision makers to make judgments that are biased toward an initially presented value. \citep[p. 35]{Furnham2011}.
\end{quote}

\noindent An important feature of these definitions is that anchoring effects involve \textit{mis-use} of information contained in the anchor: anchors must either be arbitrary \citep{Jacowitz1995} and hence unsuitable for use in future inference, or else must exert disproportionate influence \citep{Furnham2011} on future inference. It is widely known that we can also generate phenomena similar to anchoring effects, except that the anchors are informative and are used in appropriate ways. For example, manipulating the listing prices of properties changes what agents are willing to pay for them \citep{Northcraft1987}. But that is not obviously irrational, since listing prices carry information about property values. `Anchoring' in examples such as these might simply be another name for the process of learning from evidence. It is generally agreed that if there is a problem revealed by anchoring effects, it must be either that the anchors are irrelevant, or else that they exert disproportionate influence beyond their informational relevance \citep{Furnham2011,Jacowitz1995,Lieder2018}. This consensus will be important below.

\sbs{Anchoring in code generation}

Code generation tasks involve generating code from prompts. Prompts may be partial programs, English descriptions of desired functionality, or combinations of these and other inputs. Two leading code generation models are OpenAI's Codex \citep{Chen2021} and Salesforce's CodeGen \citep{Nijkamp2023}.

The HumanEval dataset is often used to assess code generation \citep{Chen2021}. HumanEval is composed of 164 programming problems. Each problem contains a three-part prompt: a function signature `def function{\_}name', an English description of the desired functionality, and several input-output pairs describing correct function behavior. Each problem is also accompanied by a canonical solution: a correct solution program generated by human programmers. 

Erik Jones and Jacob Steinhardt \citeyearpar{Jones-Steinhardt-2022} aim to find an anchoring effect in code generation by Codex and CodeGen. They do this by incorporating tempting, but incorrect solutions into `anchor' strings, then prepending anchor strings to complete HumanEval prompts. 

\begin{figure}
\begin{center}
\includegraphics[scale=0.5]{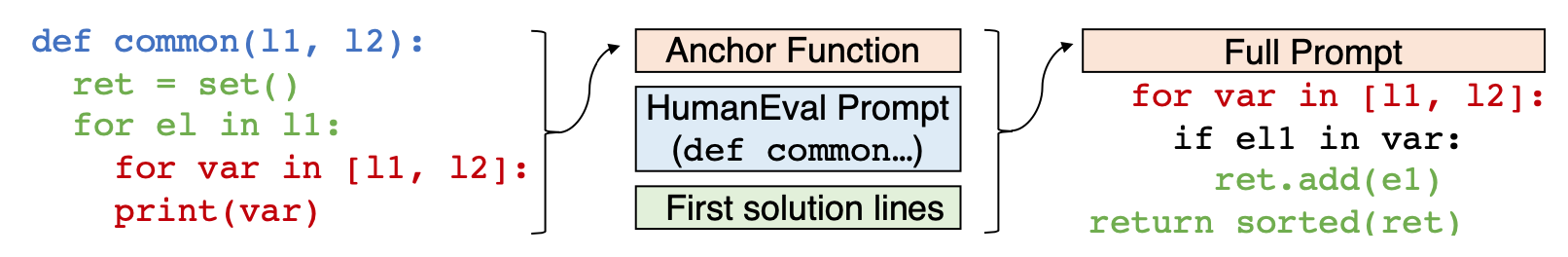}
\end{center}

\caption{Construction of anchor function and full prompt, from Jones and Steinhardt \citeyearpar{Jones-Steinhardt-2022}.}
\label{ConstructionAnchor}
\end{figure}

More concretely, Jones and Steinhardt construct anchor functions with three parts (Figure \ref{ConstructionAnchor}). The first part is the function signature, copied from the HumanEval prompt. The second part is the first $n$ lines of the canonical solution, with $n$ varied between $0$ and $8$ across prompts. The final part is a  set of `anchor lines' describing a tempting but incorrect partial solution. 

Jones and Steinhardt consider two types of anchors. \textit{Print-var} anchors instruct the program to print, rather than return, a given value:

\begin{quote}
for var in [var1, var 2]: \\
 \hspace*{10mm} print(var)
\end{quote}

\noindent \textit{Add-var anchor} lines instruct programs to sum two values:

\begin{quote}
tmp = str(var1) + str(var2) \\
 \hspace*{10mm} return tmp
\end{quote}

\noindent Complete anchor functions consist of a function signature, the first $n$ lines of the canonical solution, and the chosen anchor lines. Total prompts are constructed by prepending anchor lines to the original HumanEval prompt, consisting of a function signature, an English description of the desired functionality, and example input-output pairs (Figure \ref{ConstructionAnchor}). These are again followed by the first $n$ lines of the canonical solution, with $n$ fixed at its value in the anchor function.

Jones and Steinhardt test Codex and CodeGen across a variety of total prompts, varying the choice of anchor lines, the number $n$ of canonical solution lines, and the original prompt from HumanEval. They find a significant decrease in model accuracy, as well as an increased tendency for solutions by Codex and CodeGen to incorporate anchor lines in part or full within the resulting outputs. Jones and Steinhardt treat this finding as an anchoring effect, in which ``code models $\dots$ adjust their output towards related solutions, when these solutions are included in the prompt'' \citep{Jones-Steinhardt-2022}.

\sbs{Discussion}

The discussion in Section \ref{ResearchAnchoring} suggests three challenges for Jones and Steinhardt's anchoring experiment. First, Jones and Steinhardt sometimes talk as though they have found processes of \textit{adjustment} away from an anchor.\footnote{For example: ``Using anchoring as inspiration, we hypothesize that code generation models may adjust their output towards related solutions'' and ``We additionally find that elements of anchor function often appear in both models’ outputs, suggesting that code generation models adjust their solutions towards related solutions'' \citep{Jones-Steinhardt-2022}.} However, no evidence for any process of anchoring and adjustment has been provided. We saw in Section \ref{ResearchAnchoring} that this is important: the last time that a heuristic process of anchoring and adjustment was posited to explain anchoring effects, it turned out that this postulate was almost always wrong. This led to a clear consensus within the field that anchoring and adjustment should not be postulated without direct process-tracing evidence, which Jones and Steinhardt have not provided. This means that it is most appropriate to treat Jones and Steinhardt's finding within the broader category of anchoring effects. 

Second, the anchors provided by Jones and Steinhardt are relevant, not irrelevant. They are highly similar in content to the problem and constructed to be similar to correct solutions. This makes the anchors generally relevant to, and informative about, the problem at hand. As we have seen, most scholars concede that agents may rationally make use of relevant anchors, just as they may rationally make use of other relevant information. We may still criticize agents for \textit{over}-use of relevant anchors, just as we may criticize them for over-use of any other item of evidence, but pressing this charge requires proving over-use, which Jones and Steinhardt do not attempt to do.

Third, even if the anchors provided by Jones and Steinhardt were not in fact relevant, there would nonetheless be a legitimate presupposition of relevance. This presupposition can be grounded in two ways. The first ground for a presupposition of relevance is due to model construction. Codex and CodeGen are designed to predict likely continuations of code strings, then generate novel code according to their predictions. It is an undeniable fact that most features of code snippets are more likely to be included in the continuation if they are included in the prompt than if they are not: for example, a program that begins with a for var loop or an instruction to print variables is more likely to continue with a for var loop or an instruction to print variables. In becoming more likely to include input features in output continuations, Codex and CodeGen do no more than what they were constructed to do: take the entire input string as relevant to determining the likely continuation. 

A second way to generate a default presupposition of relevance draws on how Codex and CodeGen were trained. Both models were trained primarily on helpful and non-misleading prompts. While the models may have been exposed to natural human errors, they have not been significantly exposed to programmers trying to manipulate them into including irrelevant code in their outputs. From this, any rational agent would learn that input is likely to be non-manipulative. Codex and CodeGen do not, and should not, treat inputs as likely to be manipulative unless they are trained on manipulative examples. We could, of course, train versions of Codex and CodeGen that were designed to filter out manipulative prompts, but it is not obvious that this would be desirable unless we anticipate that many test prompts will be manipulative. 

This discussion of a default presupposition of relevance is naturally situated within the paradigm of ecological rationality \citep{Todd2012b}. This paradigm stresses that the rationality of computational processes is environment-relative. Many processes return quick, accurate, and helpful responses in some environments, but slow, inaccurate, or unhelpful responses in others. As a result, the right question to ask about a process is not how it performs in all environments, but rather how it performs in the environments where it is proposed for use. Codex and CodeGen are designed to work well on non-manipulative prompts. They do not work well on manipulative prompts, but that is not what they were designed to do. Applying Codex and CodeGen for use in hostile environments where they were never intended for use proves no more than that Codex and CodeGen should not be used, and were never intended to be used in these environments.

\sct{Framing effects: Banishing Pangloss} \label{Framing}

Bounded rationality theorists are sometimes accused of taking the Panglossian view all seeming biases and irrationalities can be explained away as nothing of the kind. Daniel Kahneman once quipped, not entirely without justification, that some theorists see only two types of errors: ``pardonable errors by subjects and unpardonable ones by psychologists'' who misinterpret them \citep[p. 349]{Kahneman1981}.\footnote{Somewhat less charitably, Keith Stanovich and Richard West contrast the `Panglossian' view that existing experiments fail to demonstrate widespread irrationality with the `meliorist' view that irrationalities are genuine and we should work to make them better \citep{Stanovich2000}.}

No theorist should be a Panglossian. It is quite likely that large language models, like humans, sometimes reason in undesirable ways. When there is clear evidence of undesirable biases in reasoning, we should do what we can to improve the situation. In this section, I want to illustrate my anti-Panglossian commitments by looking at one area where problematic biases in reasoning by large language models do seem to have been identified.

\textit{Framing effects} occur when irrelevant changes in the framing of a reasoning problem lead to substantive changes in the judgments that result from reasoning. Many authors allege framing effects in large-language models, and some of these findings may be more difficult to resist.\footnote{However, there are some negative findings. For example, John Horton \citeyearpar{Horton2023} finds that framing is largely ineffective as a manipulation in Kahneman and colleagues' \citeyearpar{Kahneman1986} classic snow shovel experiment when run on GPT-3.}

For example, Alaina Talboy and Elizabeth Fuller \citeyearpar{Talboy2023} consider a classic presentation of Tversky and Kahneman's \citeyearpar{Tversky1981} Asian disease problem. This program presents a choice between certain and risky policies, manipulating whether the outcomes of each choice are framed positively, in terms of lives saved, or negatively, in terms of those who will die. Table \ref{AsianDisease} presents the prompts used, which are formed by joining a common set of instructions together with a positive or negative framing of the policies to be considered.

 Talboy and Fuller test ChatGPT-3.5, GPT-4, and Google Bard on the Asian disease problem, finding humanlike patterns of preference change across framings. Like humans, the models opt for the safe option in the positive framing, but the risky option in the negative framing, showing risk-aversion in gains but risk-seeking in losses, even across what many would regard as equivalent problems. Extending this finding, Marcel Binz and Eric Schulz \citeyearpar{Binz2023} find human-like gain/loss framing effects in a number of classic problems: GPT-3 is loss-averse, risk-seeking in outcomes framed as losses, and risk-avoidant in outcomes framed as gains. 

\begin{table}[]
\small
\begin{tabular}{@{}ll@{}}
\multicolumn{2}{l}{\begin{tabular}[c]{@{}l@{}} \toprule \\ \bd{Common instructions:}  Imagine that the U.S. is preparing for the outbreak of an unusual \\ disease,  which is expected to kill 600  people. Two alternative programs to combat the \\ disease have been proposed. Assume that the exact scientific estimate of the consequences \\ of the two programs are as follows: \\ \end{tabular}}                                                                                                                                                                                                            \\ \midrule
\textbf{Positive frame}                                                                                                                                                                                                                                                                                & \textbf{Negative frame}                                                                                                                                                                                                                                                              \\ \midrule
\begin{tabular}[c]{@{}l@{}}If Program A is adopted, 200 people will \\ be saved.\\ \\ \\ If Program B is adopted, there is a 1/3 \\ probability that 600  people will be saved, \\ and a 2/3 probability that no people will be \\ saved.\\ \\ Which of the two programs would you favor?\end{tabular} & \begin{tabular}[c]{@{}l@{}}If Program A is adopted, 400 people will \\ die.\\ \\ \\ If Program B is adopted, there is a 1/3 \\ probability that  nobody will die, and a \\ 2/3 probability that 600 people \\ will die.\\ \\ Which of the two programs would you favor?\end{tabular} \\ \bottomrule
\end{tabular}
\normalsize
\caption{Asian disease problem, as presented in \citet{Talboy2023}.}
\label{AsianDisease}
\end{table}

Should these results be viewed as undesirable biases in need of correction? Certainly some framing effects might be defended. For example, rationalizing explanations have been offered in particular cases such as the Asian disease problem \citep{Dreisbach-2019}. And in some cases, it may be helpful to question the experimental designs that lead us to allege framing effects \citep{Lejarraga2021,Gigerenzer2018}. But even those who have wanted to defend some framing effects have not typically thought that all framing effects can be explained away, or made desirable through such means \citep{Bermudez2020}.

There may yet be some purposes for which we would like large language models to show framing effects. For example, this may enable us to use large language models as participants in laboratory studies to shed light on human reasoning \citep{Argyle2023,Aher2023,Dillion2023}. More generally, we should not exaggerate the prevalence or influence of framing effects \citep{Demaree-Cotton2016}. But in many situations, there may be reasons to find framing effects undesirable. Good reasoning responds to relevant features of situations and ignores irrelevant features. Anything else risks inconsistency, as well as a decline in the quality of judgments that are formed based on irrelevant features.

Insofar as some framing effects are undesirable, we should take two types of measures to correct them. First, programmers should explore debiasing methods to reduce the vulnerability of future models to framing effects. And second, prompt engineers \citep{Henrickson2023} should explore ways to reduce the likelihood that irrelevant prompt changes will trigger framing effects. Together, these interventions may help to improve the performance of large language models in reasoning tasks.

\section{Two lessons} \label{TwoLessons}

So far, we have discussed four types of biases alleged in large-language models: knowledge effects (Section \ref{KnowledgeEffects}), anchoring bias (Section \ref{Anchoring}), availability bias (Section \ref{Availability}), and framing effects (Section \ref{Framing}). At the beginning of this paper, I suggested that these discussions could be used to draw two lessons: a cautious optimism about model performance, and an anti-Panglossian, meliorist willingness to accept the existence of some problematic biases and work to correct them. In this section, I make the case for both lessons. Then in Section \ref{Conclusion}, I draw philosophical implications from this discussion.

\sbs{Cautious optimism} \label{CautiuosOptimism}

The cautious optimist accepts that the cognitive bias framing is useful and coherent. It makes sense to talk about large language models as showing, or failing to show, cognitive biases, and we should expect to learn something valuable about model performance by speaking in this way.

The cautious optimist reminds us of the lessons gleaned from over a half-century of discussions of bias in human cognition. In particular, she reminds us that many theorists believe that problematic biases are relatively rare, and that human cognition is often fairly rational \citep{Lieder2020,Gigerenzer2001a,Gilovich2002}. She reminds us that in the human case, many early bias accusations are now thought to depend on conceptual confusions (as in the distinction between objective and subjective availability), empirical problems (as in the difficulty of finding evidence for anchoring and adjustment), or on behavior that can be given rationalizing explanations (as in probabilistic approaches to knowledge effects). 

The cautious optimist further reminds us that many biases in human cognition are thought to arise from tradeoffs that agents face in pursuing their goals, such as a bias-variance tradeoff in predictive error \citep{Geman1992,Gigerenzer2009} or an accuracy-coherence tradeoff in reasoning \citep{ThorAccCoher}. She suggests that these tradeoffs should make us suspicious of a tendency to deem biases as irrational without further examination of how they came about. Finally, the cautious optimist reminds us that while humans can often be induced to show biases in the laboratory, biases may be relatively less common in the environments where humans ordinarily reason \citep{Todd2012b}.

The cautious optimist suggests that many of these lessons may transfer well to the study of biases in large language models. We saw, for example, that knowledge effects (Section \ref{KnowledgeEffects}) might be treated as the desirable results of good probabilistic reasoning, rather than as the undesirable results of bad logical reasoning, and that this probabilistic reconstruction is in some ways stronger in the case of machine reasoning than it is for human reasoning. We also saw that some accusations of availability bias fail to distinguish between subjective and objective availability. When they do, what is revealed is an especially benign type of objective availability conjoined with an arguably inappropriate normative standard of ignoring learned information about categories in favor of a uniform prior (Section \ref{Anchoring}). Finally, we saw that accusations of anchoring bias need conceptual clarification in terms of a particular notion of anchoring effects distinct from anchoring and adjustment; that the relevant concept of anchoring bias should be tied to a demonstration of the irrelevance of anchor information to the problem at hand; that no attempt has been made to demonstrate irrelevance; and that the anchor information is arguably both relevant, and justifiably presumed to be relevant, to the problem on which models were tested. 

From this, the cautious optimist may draw two further lessons. The first is the importance of incorporating what is already known about human bias into discussions of cognitive bias by large language models. We saw that some leading bias accusations can be softened or dissolved by applying conceptual distinctions and empirical and normative challenges familiar from the human literature, and this gives us every reason to pay greater attention to the existing literature on human cognitive bias in future studies.

The second lesson is backward-looking: Dasgupta and colleagues \citeyearpar{Dasgupta2022} suggest that insofar as machines begin to show many of the same patterns of purportedly biased cognition as humans do, this may provide supporting evidence for the claim that those biases are features, rather than bugs, in human cognition. After all, it would be a surprising coincidence if cognitive systems with very different architectures than humans were to converge on exactly the same biases, and a natural explanation for this convergence in many cases will be that there is something cognitively valuable in the bias that theories of cognition should identify and fully appreciate. I discuss this lesson in more detail in Section \ref{VindicatoryEpistemology}.

On its own, cautious optimism paints a rosy picture of bias in large language models, and to a large extent this is the picture I would like to paint. But cautious optimism must be coupled with a second reaction: anti-Panglossian meliorism.

\sbs{Anti-Panglossian meliorism}

Life is not all sun and roses. The anti-Panglossian meliorist reminds us that some biases, such as framing effects (Section \ref{Framing}) are likely to exist in large language models. While we may try to deny the existence of any particular bias, to rationalize it away, or to deny that the bias occurs often in natural environments, we should be open to the possibility that such objections will not always succeed, and may well take framing effects to be one case in which they currently fall short.

Here the anti-Panglossian meliorist agrees with the cautious optimist in accepting the usefulness of the cognitive bias framing in studying the performance of large language models. She demonstrates anti-dogmatism in taking some findings to reveal problematic biases in need of correction, and adopts a melioriative perspective which asks how our knowledge of model biases might be used to correct them and thus to improve model performance. Even the staunchest opponents of the heuristics and biases program at times show just such an anti-Panglossian meliorism, as in, for example, the use of natural frequencies to improve human probabilistic reasoning \citep{Gigerenzer1995}. The anti-Panglossian meliorist suggests that a similar spirit should be applied to some cases of machine bias.

The overall message formed by combining cautious optimism with anti-Panglossian meliorism is the following. Cognitive bias provides a novel and useful way to assess the performance of large language models. The usefulness of this approach will be improved by incorporating what is already known about cognitive bias in the human case, and when we do, current findings should be understood to paint a broadly positive picture of model performance. Nevertheless, the bias paradigm shows its teeth in areas such as framing effects, and we demonstrate genuine commitment to the usefulness of the bias framing by acknowledging the existence of a problem in such cases, then using our knowledge of how the bias is produced to create subsequent models that will produce less-biased outputs.

\section{Conclusion} \label{Conclusion}

The study of cognitive biases in language models has important descriptive and normative implications. In this concluding section, I survey two implications of cautious optimism about model biases, tempered by an appropriate dose of anti-Panglossian meliorism. 

\sbs{Bias and representative data} \label{UnrepresentativeData}

Traditional conceptions of algorithmic bias have stressed the role of unrepresentative data in driving unfair and discriminatory model behavior \citep{Fazelpour2021,Johnson2020}. Models trained primarily on white, male, western, English-speaking individuals learn to best represent and respond to the needs of those individuals. This leads to significant cross-group differences in model performance in areas as diverse as facial recognition , sentencing recommendations and medical diagnosis \citep{Buolamwini2018,Fazelpour2021}. While it is certainly not true to say that algorithmic biases should be blamed entirely on data, it is widely thought that unrepresentative data plays a leading role in driving algorithmic bias.

By contrast, to my knowledge no scholars have suggested that cognitive biases in language models emerge from unrepresentative samples of data. Certainly, nothing like this would be alleged in humans, since many cognitive biases replicate cross-culturally with sufficient frequency to cast doubt on the idea that those biases result primarily from knowledge specific to particular groups \citep{Stankov2014}. If anything, cognitive biases might be \textit{reduced} by exposure to biased samples of data. For example, there is good evidence that many, though far from all cognitive biases are less prevalent in individuals who score highly on  standard tests of cognitive ability \citep{Stanovich1999,Stanovich2000}. This might suggest that one strategy for reducing cognitive bias in language models would be to preferentially expose models to reasoning by members who perform well on tests of cognitive ability. However, most of these tests show troubling correlations along dimensions of group membership \citep{Schmidt1988}, so there may be tension between the types of data that would best reduce traditional algorithmic biases and those that would best reduce cognitive biases.

If this is right, then the need to combat unrepresentative data may be significantly greater if we are concerned with traditional conceptions of algorithmic bias than if we are concerned with cognitive bias. This means that credible evidence of widespread and undesirable cognitive biases in large language models might provide motivation for diminished focus on biases introduced by unrepresentative data. This would not be a pleasant result. By contrast, if I am right that existing evidence does not support widespread allegations of problematic cognitive biases in language models, then there will be limited impetus to reduce current focus on the role of unrepresentative data in producing harmful biases.

\sbs{Vindicatory epistemology} \label{VindicatoryEpistemology}

What leads humans to exhibit cognitive biases? In any given case, there are at least two competing descriptive explanations which can be offered, with correspondingly different normative implications.

Dual process theorists suggest that human cognition is divided into two distinctive types of processes: fast, automatic, associative and biased Type 1 processes, and slow, controlled, rule-based, normative Type 2 processes \citep{Evans2013}. On this view, bias results from the application of Type 1 processes. Biases produced in this way are likely irrational and should be corrected by application of Type 2 processes. 

Vindicatory epistemologists \citep{Thorstad2022Book-b} suggest that many biases are the result of rationalizing factors such as task demands, cognitive bounds, and the structure of the agent's environment. For example, anchoring bias may result from diminishing returns to costly processes of iterative belief adjustment \citep{Lieder2018}, and we saw in Section \ref{KnowledgeEffects} that Wason selection task behavior may result from probabilistic approaches to conditional reasoning. The vindicatory approach offers a wide array of descriptive explanations for the emergence of biases, typically resisting the dual process approach and the corresponding inference to the irrationality of observed cognitive biases \citep{Dorst2023,Icard2018,Thorstad2022Book-b}. 

Dasgupta and colleagues \citeyearpar{Dasgupta2022} conclude their discussion of knowledge effects with an interesting observation: the emergence of cognitive biases in large language models may provide some evidence for the vindicatory explanation of how those biases emerge. On the one hand, it is very difficult for dual process theorists to explain why language models should show cognitive biases, since there is no clear distinction between Type 1 and Type 2 processes in large language models. On the other hand, the emergence of similar biases in agents with very different cognitive architectures lends support to the vindicatory theorist's contention that biases emerge, not because of peculiar and irrational features of any particular cognitive architecture, but rather because of rationalizing factors such as task demands that persist across architectures. Otherwise, it would be a great mystery why similar biases should emerge across radically different agents.

This suggests that research into cognitive biases in large language models may provide an important avenue of support for vindicatory epistemology. However, this approach leaves open at least three classes of questions for future research. First, vindicatory theorists need to rule out competing explanations for the emergence of cognitive biases, such as deliberate mimicry of observed patterns of human reasoning. Second, vindicatory theorists should hope that biases are relatively stable across improvements to language models: if, as some have suggested \citep{Hagendorff2023,Horton2023}, biases are reduced in more sophisticated models, this finding might lend some support to the idea that biases result from unsophisticated reasoning processes. Finally, humans exhibit not only coarse-grained dispositions towards cognitive biases, but also fine-grained patterns of bias across prompts and tasks. The findings most friendly to vindicatory theorists would be findings in which not only coarse-grained facts, such as the presence or absence of particular biases, but also fine-grained facts about the pattern and amount of bias in particular tasks were to be similar across human agents and language models. While these findings would not settle debates about the rationality of biases in human cognition, they would represent an important step forward in our understanding of how biases come about, thought by many sides to have significant bearing on questions about the rationality of cognitive biases. 

\urlstyle{same}
 \bibliography{ManualBiblio}
 \bibliographystyle{phil_review}

\end{document}